# MAILS - Meta AI Literacy Scale: Development and Testing of an AI Literacy Questionnaire Based on Well-Founded Competency Models and Psychological Change- and Meta-Competencies

ASTRID CAROLUS[†] and MARTIN KOCH[†], Media Psychology, University of Würzburg, Germany

SAMANTHA STRAKA, Human-Computer Interaction, University of Würzburg, Germany

MARC ERICH LATOSCHIK, Human-Computer Interaction, University of Würzburg, Germany

CAROLIN WIENRICH, Psychology of intelligent interactive Systems, University of Würzburg, Germany

The goal of the present paper is to develop and validate a questionnaire to assess AI literacy. In particular, the questionnaire should be deeply grounded in the existing literature on AI literacy, should be modular (i.e., including different facets that can be used independently of each other) to be flexibly applicable in professional life depending on the goals and use cases, and should meet psychological requirements and thus includes further psychological competencies in addition to the typical facets of AIL. We derived 60 items to represent different facets of AI Literacy according to Ng and colleagues' conceptualisation of AI literacy [33] and additional 12 items to represent psychological competencies such as problem solving, learning, and emotion regulation in regard to AI. For this purpose, data were collected online from 300 German-speaking adults. The items were tested for factorial structure in confirmatory factor analyses. The result is a measurement instrument that measures AI literacy with the facets "Use & apply AI", "Understand AI", "Detect AI", and "AI Ethics" and the ability to "Create AI" as a separate construct, and "AI Self-efficacy in learning and problem solving" and "AI Self-management". This study contributes to the research on AI literacy by providing a measurement instrument relying on profound competency models. In addition, higher-order psychological competencies are included that are particularly important in the context of pervasive change through AI systems.

## 1 INTRODUCTION

It is an undeniable fact that Artificial Intelligence (AI) is coming into our daily lives. Whether for work or entertainment, interaction with AI or AI systems will become increasingly common. The broad spectrum of AI brings with it challenges

---

[†]Both authors contributed equally to this research.

Authors' addresses: Astrid Carolus, astrid.carolus@uni-wuerzburg.de; Martin Koch, martin.koch@uni-wuerzburg.de, Mediapsychology, University of Würzburg, Germany; Samantha Straka, Human-Computer Interaction, University of Würzburg, Germany, samantha.straka@uni-wuerzburg.de; Marc Erich Latoschik, Human-Computer Interaction, University of Würzburg, Germany; Carolin Wienrich, Psychology of intelligent interactive Systems, University of Würzburg, Germany, carolin.wienrich@uni-wuerzburg.de.

for understanding AI, as the underlying systems or capabilities of AI are complex and hard to grasp[49]. To find one's way in an AI-influenced world and to be able to act in a self-determined manner and participate in future developments, users need an understanding of what AI is and what it can do [8, 47]. This set of skills includes using, applying, or interacting with AI and is commonly referred to as "AI Literacy" [29].

The need of AI Literacy has already been tackled in the field of education, where efforts are made to implement AI Literacy into school curricula and develop educational approaches to teaching AI literacy [13, 19, 36, 43, 50]. However, many prospective users of AI who are already working will not benefit from AI education in schools right now. For them, their current level of AI literacy might predict if they can adapt to new technologies and if the implementation of AI-reliant workflows will be successful. As automation as well as collaboration with AI will occur in many jobs [16], it is important to also foster AI literacy in the context of work and adult education. A reliable and valid measurement instrument for AI literacy is important for the selection of suitable personnel, to identify shortages in skill and knowledge that can be addressed, and to evaluate interventions that focus on the improvement of AI literacy. So far, there are not many ways to measure AI Literacy, and many scales are bound to specific contexts (e.g., only usable in a medical setting) or are not validated [20, 46]. Moreover, there is no measurement instrument that takes into account psychological meta-competencies [47]. However, these are particularly important in the area of work and adult education since the introduction of AI systems is often accompanied by general change processes that must be mastered constructively. The present paper addresses these research gaps by conducting an empirical study on the systematic development and factorial validation of an AI literacy scale that meets psychometric requirements, is cross-contextual applicable, is embedded in the current literature on AI Literacy, and considers psychological meta-competencies.

## 2 THEORETICAL BACKGROUND

Originally, the term "literacy" referred to the basic knowledge to read and write. More modern definitions, applying a more general understanding of literacy, define it as the "ability to identify, understand, interpret, create, communicate and compute, using printed and written materials associated with varying contexts" [15]. This thus involves not only basal skills of reading and writing but also more complex thought processes of comprehension, interpretation, and creation. In recent years, the term literacy has been used for an even broader array of different competencies in regard to different domains (e.g., financial literacy, health literacy, or scientific literacy). Most subtypes of literacy, however, focus on information technology (e.g., digital literacy, media literacy, information literacy, technology literacy, information technology literacy, social media literacy [35], digital interaction literacy [8]).

In general, AI Literacy definitions differ in regard to the exact number and configuration of competencies. There are many different conceptualizations and definitions of AI Literacy: Ng and colleagues, in their review on AI literacy conceptualisations in education, postulate that these can be organized into four concepts: (1) know and understand AI, (2) use and apply AI, (3) evaluate and create AI, and (4) AI Ethics [33]. Most conceptualisations of AI literacy consider users as AI literate even if they do not have in-depth technical knowledge and are not able to develop or create AI. For example, Long & Magerko define AI Literacy as a "*set of competencies that enables individuals to critically evaluate AI technologies; communicate and collaborate effectively with AI; and use AI as a tool online, at home, and in the workplace*" [29]. Cetindamar and colleagues [10] define AI literacy as "a bundle of four core capabilities", namely *technology-related*, *work-related*, *human-machine-related*, and *learning-related* capabilities. They argue that technological capabilities will be necessary, as AI is based on technology. However, they limit these capabilities to the use of tools and data literacy and do not include in-depth programming skills to be part of AI literacy. Zhang and colleagues [52] developed a curriculum for middle schools to foster AI literacy with three components. *AI concepts* includes factual knowledge about AI and its concepts and technical details. *Ethical and societal implications* includes the ability to understand consequences of the use of AI for society, and *AI Career Future* concerns the impact of AI on future careers. However, in-depth technical knowledge about the creation of AI is not included. For their measurement instrument on AI literacy, Wang and colleagues define AI literacy to have the components *awareness*, *usage*, *evaluation*, and *ethics* but do not include the ability to develop AI applications as part of their conceptualisation [45]. As Ng and colleagues conclude [33], most conceptualisations of AI literacy parallel Bloom's taxonomy [7] regarding their general configuration of skills. Since this taxonomy is the basis for numerous competence formulations in schools and universities, the present paper also relates to it. Compared to the previous literature, we have fully incorporated all aspects of the taxonomy, including the component *creation/development of AI*, which was not previously considered.

## 2.1 Measuring AI Literacy

Several scales have been developed to measure AI literacy so far. As many articles about AI literacy stem from an educational context, many measurement instruments were developed for the evaluation of a specific intervention. Often, teaching success is measured with single-choice or open-ended knowledge tests (e.g [4, 19, 32, 37, 44, 50, 52]). The advantage of these tests is the apparently higher validity, which, however, is only given to a limited extent with open answers. A disadvantage is that they often remain close to the content of the intervention to be evaluated or to the content of the lesson. Other researchers in an educational context resort to self-assessments [11–13, 21, 22], which are easier to carry out and more objective as no interpretation of answers is necessary. Some research tends to use both options and combine [19, 52]. What all instruments used in schools have in common is that their factorial structure was not examined in large samples. A large proportion of the studies do not differentiate between different aspects of AI literacy. The majority of these questionnaires and tests might be useful for the evaluation of specific interventions. However, they are less suitable for the measurement of AI literacy in a broader spectrum of use cases for two reasons: First, they heavily depend on the specific knowledge of the tested intervention. For example, in order to assess AI literacy in general in work contexts, it should be possible to query general criteria, which can then be combined with context-specific aspects in a modular way. Second, they do not differentiate between different aspects of AI literacy. Especially for the extensive use in science and practice, it is important to differentiate between distinct facets of AIliteracy so that the questionnaire can be used economically and purposefully.

At the time of finalizing this paper, there were three published scales for measuring AI literacy that did not focus on the school but could be used for more general measurement purposes and one collection of items to measure AI literacy. Karaca and colleagues created a scale to specifically measure the *AI Readiness* of medical students in healthcare through a self-report scale [20]. However, it would be easily possible to adapt the scale to different professional fields. The Medical artificial intelligence readiness scale for Medical students (MAIRS-MS) was developed to measure the readiness of medical students for the use of artificial intelligence in their work. Exploratory and confirmatory factor analysis showed a good fit for the 27-item scale, which tends to measure the AI readiness in the four domains of "Cognition", "Ability", "Vision", and "Ethics". Another study on the development and validation of a scale to measure AI literacy was published this year [45]. The "artificial intelligence literacy scale" measures AI literacy with 12 items on the four dimensions of "Awareness", "Usage", "Evalation" and "Ethics". Again, the identified dimensions were confirmed by factor analysis. However, Wang and colleagues draw heavily on existing literature on digital literacy for their conceptualisation of AI literacy and the development of their questionnaire. Both questionnaires do not consider the current theoretical advancements regarding the conceptualisation of AI literacy [33]. The third instrument was recently presented [34]. Similar to the scale presented in this paper, the scale by Pinski and Benlian is designed to measure AI literacy of non-experts in the working context. Even though they refer to AI literacy conceptualisations [29, 33], their conceptualisation of AI literacy and structure of the questionnaire are not based on these conceptualisation. Rather, they follow conceptualisations from the field of information systems [40] to guide their scale construction.

Laupichler and colleagues present a collection of 38 items to measure AI literacy [27]. They generated an initial set of items and asked experts in the field of AI education to refine the items following the Delphi method. Their item collection targets non-expert users of AI. Items are only loosely based on a recent AI literacy framework [29] which was used as an "implicit decision support tool". Also, no factor analysis was conducted to test the factorial structure and advance AI literacy conceptualisations.

The numerous current publications show that AI literacy is an important topic, which is, however, researched in a very different and specialized way. This results in measuring instruments that, on their own, produce a coherent structure but ignore important aspects and thus have only a very limited scope of application. Established competence taxonomies like those of [7] and [33] are not consistently used as a theoretical basis for item formulation. Thus, the interpretation of latent factors also remains arbitrary. Therefore, we base our measurement instrument on the established competence taxonomies of [7] and [33].

## 2.2  A psychological perspective on the measurement of AI literacy

When measuring human competencies such as AI literacy, several options (e.g., tests, observations of behaviour, questionnaires) exist. Although tests tend to be more objective and observations tend to possess a higher validity, self-report questionnaires have a different advantage not despite but because they measure the self-perceived competence of an individual: According to the theory of planned behaviour [1] or similar theories (i.e., the social cognitive theory or extensions to the UTAUT2 [42] by different authors [2, 17, 25]), the perceived capabilities regarding a certain behavior (called perceived control or self-efficacy) are central to the intention of showing or changing a behavior. The theory of planned behavior [1] postulates that intentions and perceived behavioral control are most important to predict human voluntary behavior (such as the use of AI, or managing AI-induced changes). Intentions, in turn, are centrally predicted by attitudes towards the behavior, the subjective norm (i.e., influences from the social environment), and perceived behavioural control. Thus, perceived behavioral control takes a special role in that it influences behavior directly and indirectly (i.e., via intentions). Even though the theory of planned behavior is used mainly to predict change in health behavior and has been criticised [41], recently, it has successfully been applied to predict the intention to use AI in different domains such as agriculture [30], and human resources [3], the intention to learn AI [11] or other related new technologies [51]. Outside the theory of the planned behavior-framework, it was also shown that the subjective assessment of one's own competencies is central to the intention use AI [24, 26]. Consequently, according to an essential general psychological theory of intentional behavior, it is important to measure perceived behavioral control for the target domain of AI usage in addition to other constructs, such as the attitude towards the usage of AI or social influences. From a psychological point of view, it is, thus, reasonable to resort to a questionnaire to measure AI literacy.

However, the behavioural process does not end with the one-time formation of a behavioural intention [18]. In the further course of the use of AI, the control of action and emotion processes is necessary for the successful long-term and sustainable use of AI [18]. According to Bandura [5], several sources are central to the formation of perceived behavioral control, which he calls self-efficacy. Individuals experience higher self-efficacy when they are successful themselves when they see others being successful, when they experience no negative and high positive emotional arousal, and when they are supported by others. Especially new developments in the field of AI might hinder long-term use of AI as they can have a negative effect on the perceived behavioural control: Innovations might unsettle potential users, lead to failed behavior, and thus reduce perceived behavioral control. In addition, negative emotional states of arousal might occur, which can have further negative effects on perceived behavioural control. We suggest that other psychological competencies are central to mitigating negative effects on perceived behavioural control and thus the behavioral intention and actual use of AI. Learning and problem solving as well as emotion regulation, might be needed to mitigate the negative effects of innovations on perceived behavioral control. Especially learning and problem solving regarding AI can enable the potential AI user to keep up with current developments of AI. Learning has been considered as an important part of or addition for AI literacy before [8, 10, 13]. These competencies presumably lead to higher future use of AI by reducing failures which would lead to reduced perceived behavioral control. Additionally, emotion regulation helps to reduce negative and increase positive emotional arousal [8] and, by that, also increases perceived behavioral control. We argue that an instrument for measuring AI literacy with the aim of predicting and preparing the use of AI in a professional context should, from a psychological point of view, primarily focus on the subjective assessment of one's own competencies (i.e., behavioral control or self-efficacy). From the perspective of the theory of planned behaviour [1], behavioral control takes on a central role. In addition to subjective competence, other competencies are also important, especially to help predict and ensure the long-term use of AI. These include, in particular, the ability to learn, problem-solving skills, and emotion regulation to compensate for failures and resulting negative emotional arousal. In addition, we also consider the ability to recognise and prevent the influences of human-like voice-based AI systems [8, 48] as important. In the following, these competencies will be summarized under the term *AI Self-management*. We argue that AI Self-management is important to ensure the prolonged and sustainable use of AI and, thus, consider it to be an important part of our measurement instrument.

## 2.3 Summary and Present Work

In summary, different conceptualisations of AI literacy exist where the main focus is on the domains of basic knowledge about AI, the use of AI, and ethical aspects of AI usage [29, 33]. Fewer researchers include the development of AI as a part of AI literacy or allude to initial psychological components, as described above [8, 9, 47]. However, besides practically-oriented instruments often developed for a specific intervention study in an educational context, few measurement instruments with initial validations and theoretical foundations exist[20, 45]. Unfortunately, however, they are not based on established theoretical competency modeling, which makes the interpretation of latent factors seem arbitrary. As discussed above, the collection of self-assessments makes sense from a psychological point of view when it comes to predicting the actual use of AI. Further, the assessment of related psychological constructs we call AI Self-management (i.e., emotion regulation, problem solving, and learning in regard to AI) might help to gain important information to support long-term and sustainable use of AI at the workplace and beyond. In addition, a derivation of items from well-studied theoretical competency models seems important. With this in mind, this paper aims to meaningfully extend previous work on the conceptualization and measurement of AI literacy by developing a measurement instrument that:

(1) is deeply grounded in the existing literature on AI literacy,
(2) is modular (i.e., including different facets that can be used independently of each other) to be flexibly applicable in professional life depending on the goals,
(3) meets psychological psychometric requirements,
(4) and includes further psychological competencies in addition to the classical facets of AI Literacy.

The present study contributes to the research on the conceptualisation and measurement of AI literacy by (a) presenting a newly developed measurement instrument and (b) testing the factorial structure. The measurement instrument differs from already existing instruments in several important ways.

## 3  EMPIRICAL STUDY

Empirical data were collected online from 300 individuals with the first language German on the 3rd and 4th of November 2022 using the survey tool SoSci-Survey [28]. Participants were recruited using Prolific [14] and received a compensation of £3 for completion. The average completion time was 16:05 minutes (SD = 5:46), amounting to an average reward of £11.23/hour. In addition to the items we generated for AI Literacy following the conceptualisation by Ng and colleagues [33], the creation and development component of [7] and AI Self-management, anthropomorphism tendency, attitudes towards AI, and the willingness to use technology as well as the Big Five personality traits and demographic information were assessed mainly using standardised instruments.

### 3.1  Sample

In the present sample, the average age was 32.13 years ($SD$ = 11.66 years, ranging from 18 to 72 years). Most participants lived in Germany (77.00%) or Austria (7.00%). 145 participants considered themselves female (48.33%), while 152 participants identified as male (50.67%). 3 participants identified as diverse (1.00%). Almost one-fifth of the participants (19.67%) reported that they have never used AI in their everyday life, at school/university, or work.

### 3.2  Measures

All measures were administered online via SoSci-Survey in German. Prior to participation, the participants were informed about the general purpose of the study and gave their informed consent.

*3.2.1   AI Literacy and AI Self-management.* After reviewing the literature on AI literacy described in the theoretical background, we generated 72 items for the self-assessments in different domains of AI literacy. We focused on the four superordinate domains described by Ng and colleagues [33], namely "Know & understand AI", "Use & apply AI", "Evaluate & create AI", and "AI ethics". In addition to the items aiming at the assessment of the domains of AI literacy, we generated 12 items to measure additional constructs we deemed important for individuals working on and with AI. These constructs include (a) the ability to manage one's own emotions while interacting with AI, (b) the ability to recognise if one's decisions are influenced by AI and to stop this influence, (c) the ability to solve problems encountered while working with AI, and (d) the ability to keep up to date with current developments and inform oneself about new AI applications. These abilities have in common that they describe aspects of self-management: They include the management of one's own emotions and decisions as well as the management of problem-solving and learning processes. In the first step, for each of the domains, items were generated by one researcher. Then, the items were discussed, rephrased, rejected, and finalised by our team of researchers from the areas of human-computer interaction and psychology. From the total 84 items, 68 items were kept for subsequent analyses, including all 12 items on AI self-management and 56 items on the different facets of AI literacy. The 12 items on AI self-management and 56 items on AI Literacy were administered first. Each item included a statement about a specific ability related to one of the domains of AI literacy or AI self-management (e.g., "I can develop new AI applications."). The participants were asked to rate their own abilities using an 11-point Likert scale (0 to 10). We decided to use this scale because it can easily be understood as the certainty of being able to show a behaviour [6]. There is no scale labeling to achieve an approximate metric scale level. The only additional information the participants receive is that a value of 0 means that the ability is hardly or not at all pronounced, whereas a value of 10 means that the ability is very well or almost perfectly pronounced.

Table 1. Descriptive statistics for AI attitudes (positive and negative) and willingness to use new technology (acceptance, competence, control)

| | | *mean* | *sd* | *median* | *min* | *max* |
|---|---|---|---|---|---|---|
| AI attitude | Positive attitude | 3.60 | 0.57 | 3.67 | 1.50 | 4.92 |
| | Negative attitude | 2.74 | 0.67 | 2.75 | 1.13 | 4.50 |
| Willingness to use technology | Acceptance | 3.56 | 0.94 | 3.75 | 1.00 | 5.00 |
| | Competence | 4.83 | 0.82 | 4.29 | 1.00 | 5.00 |
| | Control | 3.85 | 0.68 | 4.00 | 1.75 | 5.00 |

*3.2.2 Attitude towards AI.* To measure attitude towards AI, we used our own German translation of the General Attitude towards Artificial Intelligence Scale (GAAIS) [39]. The scale consists of 20 items that measure positive (e.g., "There are many beneficial applications of Artificial Intelligence") and negative (e.g., "I think Artificial Intelligence is dangerous") attitudes towards AI. Participants rate their attitude towards AI on a 5-point Likert scale with the anchors strongly disagree/somewhat disagree/neutral/somewhat agree/strongly agree. In our sample, both subscales showed high internal consistencies ($\alpha$ = .88 for positive and $\alpha$ = .82 for negative attitude).

*3.2.3 Willingness to use technology.* The short scale for willingness to use technology [31] measures the acceptance of technology (i.e., "I am curious about new technical developments."), the competence (i.e., " I usually find using modern technology to be a challenge.") and the perceived control (i.e., " Whether or not I succeed in using modern technology largely depends on me.") when using new technologies with four items per scale. Participants rate the items regarding their willingness to work with technology on a 5-point Likert scale (from 1 = "not true at all" to 5 = "completely true"). The items for "competence" were recoded so that a high value indicates a high perceived competence in line with the other two scales. All three scales showed good internal consistencies (all $\alpha$ > .81).

## 3.3 Analysis

To test the factorial structure of our measurement instrument, we calculated a confirmatory factor analysis with the package lavaan for R [38] (version 0.6-12) and used robust Satorra-Bentler estimations. In the first step, we included all items on AI literacy and AI self-management. Based on the conceptual derivation / structure of the items [1], we tested if the items loaded on eight factors (i.e., Know & understand AI, Use & apply AI, Evaluate & create AI, and AI ethics and AI Problem solving, AI Learning, AI Persuasion literacy and AI Emotion regulation) that in turn loaded on two second-level factors (i.e., AI Literacy and AI Self-management; 1). For each of the lower-level factors, three to fifteen variables were included.

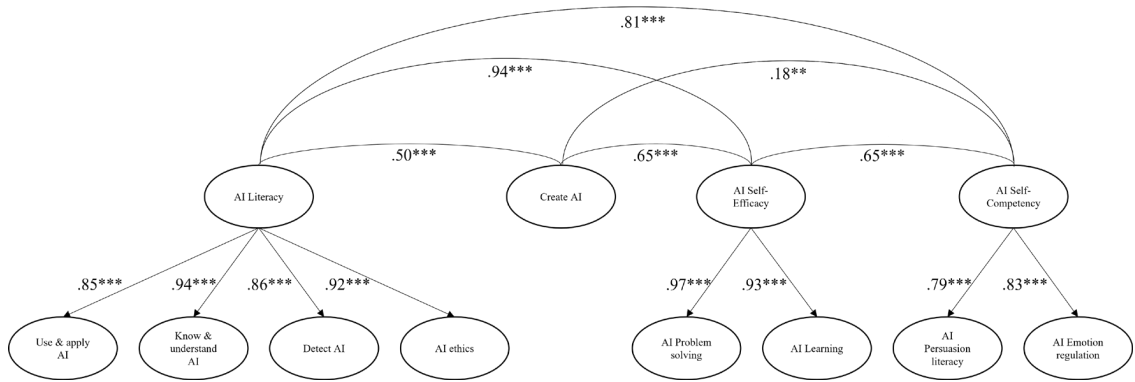

Fig. 2. Structural equation model of the modified confirmatory factor analysis. The items are omitted.

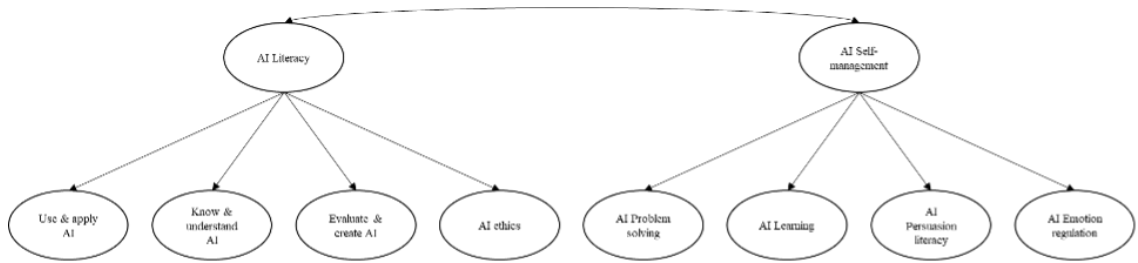

Fig. 1. Conceptual model for the confirmatory factor analysis.

Because of an insufficient model fit, in the second step, we made the following changes:

(1) We removed items that showed low factor loadings.
(2) Three items from the factor Know & understand AI were moved to a separate factor named Detect AI.
(3) The level one factor Create AI does not load on the level two factor AI Literacy.
(4) The level two factor AI Self-management is split into the factors AI Self-efficacy (includes AI Learning and AI Problem solving) and AI Self-competency (includes AI Persuasion literacy and AI Emotion regulation.

The model fit for the modified model was good. Although the $\chi^2$-test became significant ($\chi^2(513)$ = 886.87, $p < .001$), the other model fit indices showed a good model fit (CFI = .926, TLI = .920, RMSEA = .057, 95%-CI [.051, .063], SRMR = .079). All items loaded significantly on their respective factor (all $p < .001$) and all level one factors loaded significantly on their respective level two factor (all $p < .001$). All level two factors were significantly correlated (all $p < .01$). Interestingly, AI Self-efficacy and AI Self-competency showed very high correlations with AI Literacy, whereas they were still highly correlated with each other, but to a lesser extent.

In the third step, we additionally included the subscales for attitude towards AI and willingness to use technology as additional latent variables to the structural equation model. All items loaded significantly on their respective scale (all $p < .001$). The correlations of the latent variables can be seen in table 2. The model showed an acceptable model fit ($\chi^2(2035)$ = 3004.35, $p < .001$, CFI = .900, TLI = .894, RMSEA = .043, 95%-CI [.039, .047], SRMR = .069).

| | Create | AISC | AISE | Willingness to use technology | | | Attitude | |
| | | | | competence | control | acceptance | negative | positive |
| --- | --- | --- | --- | --- | --- | --- | --- | --- |
| AIL | 0.49*** | 0.86*** | 0.93*** | 0.14* | 0.20** | 0.47*** | -0.17* | 0.47*** |
| Create | | 0.24** | 0.63*** | -0.18* | -0.02 | 0.21*** | 0.10 | 0.14* |
| AISC | | | 0.72*** | 0.24** | 0.31*** | 0.36*** | -0.27** | 0.28** |
| AISE | | | | 0.06 | 0.17** | 0.48*** | -0.10 | 0.33*** |
| competence | | | | | 0.41*** | 0.42*** | -0.39*** | 0.22** |
| control | | | | | | 0.46*** | -0.27*** | 0.36*** |
| acceptance | | | | | | | -0.30*** | 0.00 |
| negative | | | | | | | | -0.43*** |

Note. * indicates $p < .05$, ** indicates $p < .01$, *** indicates $p < .001$

Self-assessed competence in the use of new technologies correlated positively only with Create AI. It did not correlate with AI Self-efficacy and only negatively with AI Literacy and AI Self-competency. All AI competencies correlated positively with the perceived control over new technology and acceptance of new technology except Create AI, which did not correlate with perceived control over new technology. A positive attitude towards AI was positively correlated with all AI competencies. A negative attitude was negatively correlated to AI Literacy and AI Self-competency but was not correlated to Create AI and AI Self-efficacy.

## 4   DISCUSSION

The aim of this paper was to develop and validate a questionnaire to assess AI literacy. In contrast to existing questionnaires, the present questionnaire aimed at: First, the questionnaire should be deeply grounded in the existing literature on AI literacy. Second, the questionnaire should be modular (i.e., including different facets that can be used independently of each other) to be flexibly applicable in professional life, depending on the goals and use cases. Third, the questionnaire should meet psychological requirements and forth, and it should include further psychological competencies in addition to the classical facets of AI literacy. For this purpose, the data of 300 German-speaking adults were analysed with exploratory and confirmatory factor analyses.

Instead of the 8 factors (4 derived from the literature [33] and 4 to measure specific psychological aspects we deemed important) that loaded on two superordinate factors (i.e., "AI Literacy" and "AI Self-management"), we found the facets "Use & Apply AI", "Know & understand AI", "Detect AI", and "AI Ethics" which loaded on a superordinate factor called "AI Literacy". Interestingly, the facet "Create AI" did not load on "AI literacy" and was only correlated to it with a $r = .5$. This result from our measurement model can be seen as support for the conceptualisations which do not include the creation of AI as a part (or dimension) of AI literacy. Rather, our research suggests that "Create AI" should be operationalized as a separate skill that is related to, but not an inherent part of, AI Literacy. This is in line with most conceptualisations of AI literacy where the development of AI is not explicitly mentioned [13, 23, 29], however, is in conflict with the conceptualisation by Ng and colleagues [33]. The structure found, where Create AI does not load on AI Literacy but is correlated to it, covers this dichotomy well, and the dimension Create AI can be measured modularly along with AI Literacy.

Concerning the domain "AI Literacy", we found further discrepancies between the first specified model and the final model. Interestingly, items for the evaluation of AI did not load on one factor with items on "Create AI" but loaded on the factor for "Know & understand AI". It seems that the ability to evaluate AI is more closely related to the general knowledge and understanding of AI than to the ability to develop AI. Presumably, according to the subjects' self-assessment, for the evaluation of AI systems, precise knowledge and understanding (i.e., "Know & understand AI") are more important than the ability to actually develop AI. This fits in with the general picture that Create AI is separate [13, 23, 29] and no part of AI Literacy. Our findings partly contradict the conceptualisation by Ng and colleagues [33]. Additionally, we found a common factor for the ability to detect AI similar to the dimensions "Awareness" (e.g., "I can distinguish between smart devices and non-smart devices." [45]). Possibly, recognizing AI does not seem to be necessarily tied to knowing and understanding AI, but also is an independent competence. This is in line with the conceptualisation by Long and Magerko [29], who included the ability to recognize AI in their framework. However, as expected, we found the other factors with the expected items to load on the superordinate factor "AI Literacy".

In place of one domain called "AI Self-management" which includes the factors "AI Problem solving", "AI Learning", "AI Persuasion literacy", and "AI Emotion regulation", we found two superordinate domains we called "AI Self-efficacy" (including the factors "AI Problem solving" and "AI Learning") and "AI Self-competency" (including "AI Persuasion literacy" and "AI Emotion regulation". Possibly, problem solving and learning are competencies that are primarily aimed at managing information and information processing, while persuasion literacy and emotion regulation also focus on the management of information, albeit with greater personal value (own decisions and emotions). We follow the opinion that these factors are important for the prolonged and sustainable use of AI tools [8–10, 13, 47]. A clear correlation emerged for "AI Self-efficacy" and "AI Self-competency", although the subordinate factors do not load on a common factor. It is possible that similar cognitive processes are necessary for the management of processes that are more concerned with the processing of information ("AI Self-efficacy") or emotions and decisions ("AI Self-competency"), leading to a high correlation among both domains. However, enough uniqueness exists in both constructs. So far, there is some conceptual thinking to include such aspects [8–10, 13, 47], but no measurement tool yet. Our measurement instrument, therefore, provides new value and an important contribution to existing considerations. This is the first step to regarding the individual differences in such a general measuring instrument [48].

There are high correlations between the constructs of our questionnaire (i.e., "AI Literacy", "AI Self-competency", "Create AI", and "AI Self-efficacy"), but still enough differentiation that they can be understood as unique constructs. All superordinate dimensions positively correlate with a positive attitude towards AI. "AI Literacy" and "AI Self-competency" negatively correlate to a negative attitude towards AI while "Create AI" and "AI Self-efficacy" do not. Presumably, individuals with a higher positive attitude also deal with the topic more often and are therefore more competent. Similar is true for negative attitude (but only for some of the constructs), with lower negative attitude going hand in hand with higher literacy and self-competence. In a similar way, the three dimensions of the willingness to use technology are positively related to most of the dimensions of our questionnaire. These correlative findings suggest that there are relations between competencies related to AI and attitudes towards AI as well as the willingness to use new technology. It appears that individuals with higher positive attitudes, lower negative attitudes, and more willingness to use technology are also more likely to consider themselves competent [1, 46, 48]. One reason may be that these individuals are also more likely to use AI and thus become more competent.

## 4.1   Limitations and Future Work

Our sample was collected online and is specific to German-speaking individuals who live in Germany or Austria. Also, it was not possible to use an already existing and validated instrument on AI literacy to validate our questionnaire, as would be the gold standard for scale development. Although other instruments exist, non have been validated with external criteria so far.

Future research should aim to test the factorial structure we identified in independent samples. Even though we chose confirmatory factor analysis to test our models, we made changes to the model, thus, it is necessary to consider our approach exploratory. The most important next step, however, is the validation of our questionnaire. This could be done by either correlating our questionnaire results with results of tests, tasks, or the evaluations of individual AI literacy by an expert, or by correlating it to a validated questionnaire that might be published in the future. Similarly, other external criteria could be used: We could test if the instrument is capable of detecting expected group differences or finding change after interventions/in the course of professional studies. Additionally, a comparison with the other existing instruments regarding their predictive validity (e.g., use them to predict the quality of future AI-related behavior) would be interesting to identify the worth of the additional scales we included.

## 4.2  Conclusion

The goal of the current study was to develop and validate a questionnaire to measure AI literacy and also include other psychological competencies that might be helpful in predicting the prolonged and sustainable use of AI. We based our developed items on the existing literature on AI literacy and psychological competencies. Overall, we find the factors "Use & Apply AI", "Know & understand AI", and "AI Ethics" [33] with the addition of "Detect AI". The factor "Evaluate & Create AI" was not found. While the items on the evaluation of AI loaded on "Know & understand AI", the items on the creation of AI formed an own factor that cannot be seen as an inherent part of AI literacy. Instead of finding one superordinate factor for the psychological competencies related to AI, we found two (i.e., "AI Self-efficacy" and AI "Self-competency"). We mainly found positive relations for our questionnaires' dimensions with attitudes towards AI and willingness to use technology. "Create AI" is a notable special case in that its correlations to attitudes and willingness are comparatively low compared to the other dimensions of our scale. Additional research will be needed to relate our measure to other valid measures of AI literacy and compare their predictive validity. The current study contributes to the existing research by providing a measurement instrument for AI literacy that is based on the current literature on AI literacy, includes important psychological constructs, and has a valid factorial structure. The instrument will be useful for researchers, practitioners, and educators who plan to measure AI Literacy and related constructs.

## A A STRUCTURE OF THE QUESTIONNAIRE

|                        | Sources                                      | Item                                                                                          |
|------------------------|----------------------------------------------|-----------------------------------------------------------------------------------------------|
| **AI Literacy**        |                                              |                                                                                               |
| Use & Apply AI         | Ng et al., 2022                              | I can operate AI applications in everyday life.                                                |
|                        |                                              | I can use AI applications to make my everyday life easier.                                     |
|                        |                                              | I can use artificial intelligence meaningfully to achieve my everyday goals.                   |
|                        |                                              | In everyday life, I can interact with AI in a way that makes my tasks easier.                  |
|                        |                                              | In everyday life, I can work together gainfully with an artificial intelligence.              |
|                        |                                              | I can communicate gainfully with artificial intelligence in everyday life.                    |
| Know & Understand AI   | Ng et al., 2022                              | I know the most important concepts of the topic "artificial intelligence".                     |
|                        |                                              | I know definitions of artificial intelligence.                                                |
|                        |                                              | I can assess what the limitations and opportunities of using an AI are.                        |
|                        |                                              | I can assess what advantages and disadvantages the use of an artificial intelligence entails. |
|                        |                                              | I can think of new uses for AI.                                                               |
|                        |                                              | I can imagine possible future uses of AI.                                                     |
| Detect AI              | Long & Magerko, 2020                         | I can tell if I am dealing with an application based on artificial intelligence.               |
|                        | Wang et al., 2022                            | I can distinguish devices that use AI from devices that do not.                                |
|                        |                                              | I can distinguish if I interact with an AI or a "real human".                                  |
| AI Ethics              | Ng et al., 2022                              | I can weigh the consequences of using AI for society.                                          |
|                        |                                              | I can incorporate ethical considerations when deciding whether to use data provided by an AI. |
|                        |                                              | I can analyze AI-based applications for their ethical implications.                            |
| **Create AI**          |                                              |                                                                                               |
|                        | Ng et al., 2022                              | I can design new AI applications.                                                              |
|                        |                                              | I can program new applications in the field of "artificial intelligence".                      |
|                        |                                              | I can develop new AI applications.                                                            |
|                        |                                              | I can select useful tools (e.g., frameworks, programming languages) to program an AI.         |
| **AI Self-Efficacy**   |                                              |                                                                                               |
| AI Problem solving     | Ajzen, 1985                                  | I can rely on my skills in difficult situations when using AI.                                 |
|                        |                                              | I can handle most problems in dealing with artificial intelligence well on my own.            |
|                        |                                              | I can also usually solve strenuous and complicated tasks when working with artificial intelligence well. |
| Learning               | Carolus et al., 2022                         | I can keep up with the latest innovations in AI applications.                                  |
|                        | Cetindamar et al., 2022                      | Despite the rapid changes in the field of artificial intelligence, I can always keep up to date. |
|                        | Dai et al., 2020                             | Although there are often new AI applications, I manage to always be "up-to date".              |
| **AI Self-Competency** |                                              |                                                                                               |
| AI Persuasion literacy | Carolus et al., 2022                         | I don't let AI influence my in my everyday decisions.                                          |
|                        |                                              | I can prevent an AI from influencing me in my everyday decisions.                              |
|  |                                              | I realise if artificial intelligence is influencing me in my everyday decisions.              |
| AI Emotion regulation  | Carolus et al., 2022                         | I keep control over feelings like frustration and anxiety while doing everyday things with AI. |
|                        |                                              | I can handle it when everyday interactions with AI frustrate or frighten me.                   |
|                        |                                              | I can control my euphoria that arises when I use artificial intelligence for everyday purposes. |